\documentclass[runningheads]{llncs}
\usepackage[T1]{fontenc}
\usepackage{graphicx}
\usepackage{amsmath}
\usepackage{amssymb}
\usepackage{subcaption}
\usepackage{hyperref}

\usepackage{float}

\begin{document}

\title{Analysis of Bluffing by DQN and CFR in Leduc Hold'em Poker}

\author{
  Tarik Začiragić \and
  Aske Plaat \and
  K. Joost Batenburg
}

\institute{
  LIACS, Leiden University, The Netherlands \\
  \email{tarikzacir@gmail.com}
}

\maketitle

% abstract and references should fit on one page
\begin{abstract}
In the game of poker, being unpredictable, or bluffing, is an essential skill. When humans play poker, they bluff. However, most works on computer-poker focus on performance metrics such as win rates, while bluffing is overlooked. In this paper we study whether two popular algorithms, DQN (based on reinforcement learning) and CFR (based on game theory), exhibit bluffing behavior in Leduc Hold'em,
%when they are trained on each other simultaneously. 
a simplified version of poker.  We designed an experiment where we let the DQN and CFR agent play against each other while we log their actions. We  find that both DQN and CFR exhibit bluffing behavior, but they do so in different ways. Although both attempt to perform  bluffs at different rates, the percentage of successful bluffs (where the opponent folds) is roughly the same. This suggests that bluffing is an essential aspect of the game, not of the algorithm. Future work should look at different bluffing styles and at the full game of poker. Code at \url{https://github.com/TarikZ03/Bluffing-by-DQN-and-CFR-in-Leduc-Hold-em-Poker-Codebase}.
\keywords{Poker \and Leduc Hold'em \and Bluffing \and Reinforcement Learning \and Game Theory \and Counterfactual Regret Minimization (CFR) \and Deep Q-Networks (DQN) \and Imperfect-information games}
\end{abstract}

\section{Introduction}\label{introduction}

Bluffing is a key feature of imperfect-information games, where players must act unpredictably under uncertainty about opponents resources, intentions, or private cards. In poker, decisions depend not only on actual hand strength but also on beliefs about how opponents perceive and react. Successfully executing a bluff involves both masking weakness and exploiting the opponents uncertainty. These are skills that are traditionally associated with human intuition and behavior \cite{vonneumann1944theory}. However, with recent advances in artificial intelligence and reinforcement learning, the question arises: Can algorithmic agents also learn to bluff? 

This paper investigates the emergence and characteristics of bluffing behavior in two widely studied AI algorithms, namely, Deep Q-Networks (DQN) \cite{mnih2015human} and Counterfactual Regret Minimization (CFR) \cite{10.5555/2981562.2981779}. The two algorithms are based on opposite principles: reinforcement learning is reactive, training its policy on feedback, where game theoretic algorithms are based on (forward looking) first principles analysis of the game. In order to facilitate extensive experimentation, we use the simplified poker environment of Leduc Hold'em, a version that includes essential poker elements such as hidden information, betting, and deception. Agents were trained against each other simultaneously, allowing mutual adaptation, and later they were evaluated against each other. The information that was logged from the games were then used for bluffing analysis.
%using two approaches: heuristic threshold rules and statistical-based detection.

By studying bluffing tendencies and reactions of DQN and CFR, this paper sheds light on how artificial agents handle uncertainty and deception.

%\subsection{Contribution}\label{Introduction Contribution}

The key contributions of this work are as follows:
\begin{itemize}
%    \item Simultaneous Cross-Paradigm Training: Developed a training framework that enables reinforcement-learning based (DQN) and  game-theoretic (CFR) agents to be trained against each other in real time, allowing for mutual adaptation and strategy co-evolution. 
    \item Using a Threshold-Based and a Statistical-Based Bluff Detection Framework we were able to define and identify bluffing attempts of both agents.
%    \item Statistical Bluff Detection Framework: Designed a probabilistic detection system that uses belief distribution analysis with statistical classification to capture deceptive behavior. 
%    \item Extended Game Complexity: Implemented a 52 card version of Leduc Hold'em to allow for an expanded strategic space which increases the number of possible bluffing scenarios. 
    \item Both the game theoretic algorithm and the reinforcement learning algorithm exhibit bluffing. However, reinforcement learning (DQN) and game-theoretic (CFR) approaches showed different bluffing strategies: DQN attempts to bluff more conservatively, but since its bluffs are more successful, overall performance is comparable. The response to perceived bluffs shows that both agents act in a very similar way even though they belong to different paradigms. 
\end{itemize}

\section{Related Work}\label{Background}

%\subsection{Achievements in CFR, DQN and AI Poker}\label{Background Big Achievements}

%Over the years, 
Significant progress has been made in the application of AI to games of strategy, and in particular, to poker. Counterfactual Regret Minimization (CFR) has been a foundational algorithm for solving imperfect-information games and has shown remarkable success \cite{10.5555/2981562.2981779,moravcik2017deepstack}. 
%In 2017, DeepStack reached expert-level performance in heads-up no-limit Texas Hold'em using an approach  combining CFR with deep learning
%, using neural networks to estimate counterfactual values in future game states during real-time play. This depth-limited continual re-solving strategy allowed DeepStack to reach expert-level performance in heads-up no-limit Texas Hold'em, demonstrating the feasibility of integrating machine learning with game-theoretic reasoning 
%\cite{}, and i
In 2019 Pluribus
%an AI system built on CFR variants and abstraction techniques, 
achieved superhuman performance in six-player no-limit Texas Hold'em, marking a milestone in multiplayer game AI \cite{brown2019superhuman,plaat2020learning}. 
%It consistently defeated professional poker players without relying on prior human data or strategies. 
%
Simultaneously, Deep Q-Networks (DQN) revolutionized the field of reinforcement learning by combining deep learning with Q-learning to play Atari 2600 games at human level of performance
%. Introduced by Mnih et al. (2015), DQN achieved human-level control on Atari 2600 games, proving that deep neural networks could learn effective policies directly from high-dimensional inputs 
\cite{mnih2015human}. 
%
%Beyond poker, multi-agent reinforcement learning has also demonstrated superhuman achievements. In 2019, DeepMind's AlphaStar reached grandmaster level in StarCraft II, %a complex real-time strategy game,
%showcasing the ability of %cooperative and competitive 
%RL agents to master long-term planning in highly stochastic environments \cite{vinyals2019grandmaster}. 

%These milestones highlight the capability of algorithmic agents to match, and in many cases, to surpass human expertise in strategically complex domains. 

\subsection{Leduc Hold'em}\label{Definitions Leduc Hold'em}

Leduc Hold'em is a simplified variant of poker which is commonly used in game theory research because of its small state  and action space, making it computationally tractable while retaining essential features of the game
%. It was  introduced  in the 2005 paper "Bayes' Bluff: Opponent Modeling in Poker" 
\cite{southey2005bayes}.
%as a minimal yet representative poker environment for studying decision-making in imperfect-information games. 

The classic version of Leduc uses a six-card deck (two copies each of King, Queen, and Jack). In this work, however, we extend the game to a full 52-card deck with 13 ranks and four suits, creating richer opportunities for deception. The game is played heads-up over two betting rounds. Each player bets first, with the first player posting one chip and the other player two chips. At the start, each player receives one private card, hidden from the opponent. Betting follows a fixed-limit structure and players may fold, call, or raise, but the number of raises per round is capped and bet sizes are predetermined.

After the first betting round, a single public card is revealed, followed by a second and final betting round. If neither player folds during the course of the game, a showdown occurs at the end of the second betting round with the winner determined by the following rules: pairs beat high cards and suits serve as tie-breakers (A > K > ... > 2; Spades > Hearts > Diamonds > Clubs). If one player folds, the other immediately wins the pot.

Despite its simplicity, Leduc Hold'em preserves several core strategic features of full-scale poker: hidden information, shifting hand strength after the public card, and the potential for bluffing and deception.

\subsection{Bluffing}\label{Definitions Bluffing}
%In imperfect information games such as Poker and Rock-Paper-Scissors, the optimal strategy is a mixed strategy---of deterministic and non-deterministic actions
%---since any fully deterministic strategy will be predictable and thus exploitable by the opponent 
%\cite{fudenberg1991game}. In order not to lose, a player should have an unpredictable (non-deterministic) strategy, since opponents would exploit any predictable reactions. 
%In Rock-Paper-Scissors the mixed strategy can be achieved through randomizing actions, in Poker we bluff.

Bluffing is one of the most iconic aspects of poker. It involves betting high with low cards: a player intentionally misrepresenting the strength of their hand by typically betting or raising with weak hands in order to convince opponents to fold their stronger hands. The effectiveness of a bluff relies on the opponents perception of the players behavior, previous actions, betting patterns, and psychology. The goal is not just to win individual hands, but to remain unreadable over time \cite{sklansky1987theory}.

A successful bluff relies not only on bold action, but also on an understanding of how one's behavior may be perceived by others. As such, bluffing is often regarded as a test of psychological insight, timing and the ability to manipulate expectations under uncertainty \cite{sklansky1987theory}.  
%In regular real-life poker played between humans, players often rely on reading behavioral cues such as body language, facial expressions, gaze, etc. Skilled players may exploit opponents who bluff too frequently or not enough. 
In contrast, AI agents operating under purely algorithmic strategies bluff as a byproduct of optimal play, rather than psychological manipulation. 

From a game-theoretic perspective, bluffing is essential for maintaining an unpredictable strategy and ensuring that a player is difficult to exploit. Without bluffing, a player that  only bets strong hands becomes exploitable. By bluffing appropriately, a player makes it harder for opponents to infer their hand strength, which in turn forces them to make a decision under uncertainty \cite{chen2006mathematics}. 

%A significant contribution to the analysis of bluffing behavior in poker comes from Southey et al. in Bayes Bluff: Opponent Modeling in Poker (2005). The authors propose a 
Bluffing has been analyzed theoretically using
Bayesian probabilistic model that separates the uncertainty of the game dynamic from the uncertainty in an opponents strategy \cite{southey2005bayes}. Their framework enables inference over opponent strategies based on observed actions, allowing the agent to compute a posterior distribution over possible opponent policies. %While it is not explicitly focused on bluff detection and analysis, their work introduces several methods (Bayesian Best Response, MAP response, and Thomspon sampling) that adapt agent behavior based on observed opponent tendencies. 
Testing their methods in both Leduc and Texas Hold'em, they demonstrate that even with limited data, Bayesian agents can learn to exploit opponents effectively including learning to respond to deceptive strategies like bluffing. 
%Their use of both independent Dirichlet priors and informed priors based on expert knowledge provides a strong precedent for modeling strategy uncertainty, which closely aligns with the goals of analyzing and interpreting bluffing behavior in learned agents. 

\subsection{Counterfactual Regret Minimization (CFR)}\label{Definitions CFR}

Counterfactual Regret Minimization (CFR) is a prominent iterative algorithm for computing approximate Nash equilibria in extensive-form games with imperfect information %such as poker 
\cite{10.5555/2981562.2981779}. %In such games, players do not have full knowledge of the game state, which makes it necessary to reason over information sets which are collections of decision nodes that are indistinguishable from the players perspective.
CFR works by simulating repeated plays of the game. CFR tracks for each decision point in the game %(information set), 
how much a player "regrets" not having chosen each available action in the past. These regrets are computed using counterfactual values---expected outcomes assuming the player had taken a different action, while the rest of the game proceeds as before \cite{10.5555/2981562.2981779}. 

%In each iteration, the algorithm performs a traversal of the game tree, during which, for the CFR agent it 1. Computes the counterfactual values of each action at each information set, 2. Updates the cumulative regret for not having taken each action, and 3. Updates the strategy at each information set using regret matching. With regret matching, actions with higher positive cumulative regrets are assigned higher probabilities and if all regrets are non-positive, the algorithm defaults to a uniform strategy.
%
%Rather than relying solely on the strategy from a single iteration, 
CFR maintains and outputs an average strategy over all iterations, which significantly improves convergence properties \cite{10.5555/2981562.2981779}. 

\subsection{Deep Q-Networks (DQN)}\label{Definitions DQN}
Deep Q-Networks (DQN) are a class of model-free reinforcement learning algorithms that approximate the optimal action-value function using deep neural networks \cite{sutton2018reinforcement,plaat2020learning,mnih2015human}. %It is a combination of the theoretical framework of Q-Learning and the function approximation capabilities of deep learning \cite{mnih2015human}.  

Q-Learning aims to learn the optimal Q-function $Q^*(s,a)$ that estimates the expected cumulative reward of taking an action $a$ in state $s$ and following the optimal policy $\pi^*$ thereafter \cite{watkins1992qlearning,wang2018assessing}. While tabular Q-learning stores values for each state-action pair, this quickly becomes infeasible in complex environments. DQN addresses this by using a neural network to estimate the Q-value function $Q(s, a;\theta)$, where $\theta$ represents the parameters (weights and biases) of the neural network \cite{mnih2015human}. The network takes a state $s$ as input and outputs an estimated Q-value for each possible action, enabling generalization to unseen states. 

%DQN has three components. The main Q-network approximates the action-value function $Q(s, a; \theta)$ using a neural network that maps states to action values. A separate target network $Q(s, a; \theta^-)$ is updated less frequently to provide stable targets for training and to reduce oscillations. Finally, the experience replay buffer stores past transitions $(s, a, r, s')$ allowing the agent to train on randomly sampled minibatches, which breaks correlations between consecutive experiences and improves data efficiency.

%The training loop alternates between collecting experience, sampling minibatches from the experience replay buffer, computing target Q-values via the target network, and updating the main Q-network. Over time, this process converges toward policies that maximize expected cumulative reward.

To date, no prior work has directly compared nor analyzed the bluffing behavior of agents trained with CFR (Counterfactual Regret Minimization) and DQN (Deep Q-Networks) in the Leduc Hold'em environment. Even in larger-scale domains like Texas Hold'em, existing studies have focused on win rates and exploitability, but not on the qualitative nature of strategies such as bluffing or deception. We address this gap by providing a comparative analysis of bluffing strategies in CFR and DQN agents. The goal is 
%not just to measure performance, but 
to interpret whether and how bluffing manifests itself in these %fundamentally 
different algorithmic approaches.

\section{Methods}\label{Methodology}

%\subsection{Environment Setup}\label{Methodology Environment Setup}

All experiments were conducted using our extended version of Leduc Hold'em, building upon the Python RLCard framework \cite{DBLP:journals/corr/abs-1910-04376}. %While maintaining the fundamental structure and strategic characteristics of Leduc Hold'em, our implementation increases the strategic complexity and provides a more realistic poker environment for bluffing analysis. We expand from using a 6-card deck to a full 52-card deck.
The RLCard implementation of Leduc Hold'em is structured and was modified in the following ways:
\begin{itemize}
    \item LeducholdemGame: The main game class which allows gameplay. It manages players, dealer, rounds, and transitions between game states. It was modified to handle a 52-card deck initialization and an expanded state space.
    \item Player: Represents each agent in the game, stores private hand information, chip count and betting status. It remained unchanged.
    \item Dealer: Manages card dealing, including assigning private hands and revealing the public card. It was modified to manage a full 52-card deck.
    \item Round: Controls the betting rounds, including turn order, allowed actions, raise limits, and round transitions.
    \item Judger: Determines the winner at the end of the game based on hand strength. It was modified to handle the 52-card deterministic hand evaluation. No ties were achieved by using suits as tie-breakers.
\end{itemize}
The environment uses the following fixed parameters described in Table \ref{Table Environment parameters}:
\begin{table}
\centering
\caption{Leduc Hold'em Environment Parameters}
\label{Table Environment parameters}
\begin{tabular}{|l|l|}
\hline
\textbf{Property} & \textbf{Description} \\
\hline
Number of players & 2 (fixed) \\
\hline
Deck & 52 cards (4 suits of each of the 13 ranks) \\
\hline
Blinds & Small blind: 1 chip bet; Big blind: 2 chip bet \\
\hline
Raise amounts & Pre-flop: 2 chip bet; Post-flop: 4 chip bet \\
\hline
Raise cap & Maximum of 2 raises per player per round \\
\hline
\end{tabular}

\end{table}

\subsection{Implementation of DQN and CFR}\label{Methodology DQN}

RLCard provides a modular implementation of DQN which is tailored for training agents in imperfect-information card games like poker \cite{DBLP:journals/corr/abs-1910-04376}. The PyTorch-based version closely follows the original DQN algorithm \cite{mnih2015human} with several practical adaptations to support batch training, evaluation, and integration into the RLCard framework. 

%At the high-level, the RLCard DQN agent consists of a Q-Network, Target Network, and an Experience Replay Buffer. Both networks are built using multilayer perceptrons with configurable depth. Particularly, this implementation of DQN uses Double DQN (DDQN). The reason for this is that one of the key limitations of the original DQN algorithm is the tendency to overestimate Q-values during training \cite{mnih2015human}. This overestimation arises because the same Q-network is used to both select the action with the highest predicted value and to evaluate that actions value. This can lead to overly optimistic estimates, especially in noisy or stochastic environments. Double DQN (DDQN), proposed by van Hasselt et al. (2016), addresses this issue by decoupling the action selection from the action evaluation \cite{van2016deep}. In DDQN, the main Q-network is used to select the best action for the next state, but the target network is used to evaluate the value of that action. This separation reduces bias and leads to more stable learning, especially in environments with high variance in rewards or long horizons. 

%The agent follows an $\epsilon$-greedy policy during training, where it selects a random legal action with probability $\epsilon$ to encourage exploration, and with probability $1 - \epsilon$, it choose the action with the highest predicted Q-value. Epsilon decays linearly over a fixed number of steps which shifts the agent from being highly explorative early on in the training process to being exploitative later on. 

Training is performed at regular intervals during environment interaction. At each training step, a batch of transitions is sampled, and the Q-network is updated to minimize the MSE (Mean-Squared Error) between predicted Q-values and target values. To stabilize learning, the weights of the target network are periodically updated to match those of the main Q-network.
Since the implementation is meant to be used with card games, it ensures action legality by masking out invalid actions. The predicted Q-values for illegal actions are set to $-\infty$ which ensures that they are never selected during action choice. 

%\subsection{Implementation of CFR}\label{Methodology CFR}

To support asymmetric training scenarios and interaction with arbitrary opponent policies, we implemented a custom version of Counterfactual Regret Minimization (CFR), based on RLCards version, that allows CFR to train against a separate, externally defined opponent. In our implementation, CFR updates its strategy by traversing the game tree while treating the opponent's actions as fixed and externally provided by a callable policy interface.%This departs from the standard CFR implementation, which relies on self-play by simulating both players using the same CFR algorithm. However, it still closely follows the standard CFR implementation \cite{10.5555/2981562.2981779}. In our implementation, CFR updates its strategy by traversing the game tree while treating the opponent's actions as fixed and externally provided by a callable policy interface. This enables the agent to train against any opponent. 

%During each iteration, the CFR agent performs a recursive traversal of the game tree. When it is the CFR agent’s turn to act, it selects actions according to its current policy and computes counterfactual utilities. When it is the opponent’s turn, actions are selected using the opponent’s current policy via a function call. This distinction allows CFR to compute regrets relative to a dynamic opponent rather than simulating both sides itself. After traversal, regrets are updated based on the difference between actual action utilities and the expected utility, and the average policy is incrementally refined using reach probabilities. 

%Information sets are represented by string-encoded observation vectors, allowing generalization across structurally identical states. Legal action masking is applied to ensure only valid actions are considered in each state, which is a crucial feature in imperfect information games like Leduc Hold’em. 

For evaluation and gameplay, the agent selects actions from its average policy, which converges to a more stable and robust strategy over time. %The internal data structures, namely, regrets, policies, and iteration count are serialized to save at regular intervals to support checkpointing and reproducibility. 
This custom CFR implementation provides a flexible framework for studying interactions with diverse opponents, making it suitable for analyzing response strategies, exploitability, and strategic adaptation in multi-agent environments.

\subsection{Simultaneous Training and Evaluation of CFR and DQN}\label{Methodology Simulatenous Training}

Training DQN and CFR against each other enables both agents to co-evolve their strategies in real time, continuously adapting to the opponent’s changing policy. 

During the training phase the agents played 100 000 games of Leduc Hold'em against each other. Each training episode consisted of both agents interacting in a shared environment, with CFR always assigned as Player 1 and DQN as Player 0. 
%At the start of each training episode (game), CFR performs 10 recursive tree traversal iterations, updating its strategy against the current DQN policy that during those iterations does not change. Then both agents play against each other in real time, where CFR samples from its average policy and DQN uses its epsilon-greedy action selection. 
Before each training episode, CFR performs 10 recursive tree traversal iterations. After a single episode, DQN updates its Q-network using trajectories (data) collected during the episode which allows DQNs strategy to evolve. This creates a continuous feedback loop where both agents adapt to each others evolving strategies. This process continues for 100 000 episode where in each episode 1 game of Leduc Hold'em is played. 

%Training progress is monitored through periodic evaluation every 5000 episodes, where 2000 independent evaluation games are played using the agents current learned policies where the policies do not change. From these games we derive the win rates of both agents which allows us to track whether the agents are getting stronger over time, whether one agent is dominating the other and whether the win rates are stabilizing which would indicate convergence. 

The DQN agent uses the following hyperparameters: 

\begin{table}
\centering
\caption{DQN Hyperparameters}
\label{Table DQN Hyperparameters}
\begin{tabular}{|l|l|l|}
\hline
\textbf{Hyperparameter} & \textbf{Value} & \textbf{Description} \\
\hline
Network Architecture & [256, 256] & Hidden layer sizes (fully connected) \\
\hline
Learning Rate & 0.00005 & Adam optimizer learning rate \\
\hline
Batch Size & 64 & Mini-batch size for training \\
\hline
Epsilon Start & 1.0 & Initial exploration probability \\
\hline
Epsilon End & 0.05 & Final exploration probability \\
\hline
Epsilon Decay Steps & 10,000 & Steps for $\epsilon$-greedy exploration decay \\
\hline
Replay Memory Size & 20,000 & Maximum replay buffer capacity \\
\hline
Replay Memory Init Size & 500 & Initial experiences before training \\
\hline
Update Target Estimator Every & 1,000 & Steps between target network updates \\
\hline
Discount Factor ($\gamma$) & 0.99 & Future reward discount factor \\
\hline
Train Every & 1 & Training frequency (every N steps) \\
\hline
Weight Initialization & Xavier Uniform & Neural network weight initialization \\
\hline
Loss Function & MSE & Mean squared error loss \\
\hline
Optimizer & Adam & Gradient descent optimizer \\
\hline
Activation Function & Tanh & Hidden layer activation function \\
\hline
Batch Normalization & BatchNorm1d & Input layer normalization \\
\hline
\end{tabular}

\end{table}

%Some hyperparameters from Table 2 are the default values of RLCards DQN agent, however, some hyperparameters such as network architecture size, learning rate, batch size, etc. have been modified to larger/smaller values in order to accompany the size of the training process. It was not feasible to run a hyperparameters sweep/optimization as the whole training process takes about an hour and for a meaningful sweep it would take days. 

After 100 000 training episodes, the trained models were saved, and an evaluation phase was conducted where another 100 000 evaluation games were played with the agents using their learned policies. %These games are logged for the analysis of bluffing behavior. We log the game id, player id, public hand, private hand, betting information and action information.

\subsection{Bluff Detection}\label{Methodology Bluff detection}

First we define the threshold-based bluff detector. To assess whether bluffing occurred, we %define a simple but effective heuristic that 
classify large raise actions as bluff attempts based on private hand strength and public card information. %The core idea is that a 
%A bluff is attempted when an agent aggressively raises despite holding a weak hand. %The action would not be justified based on hand strength alone. 
%
%The agents private hand is evaluated using the card\_score() function which assigns a base score to the hand. 
The hand score formula used is 
\begin{equation}
\text{$Hand Score$} =
\begin{cases}
(R_{pc} \times 4) + S_{pc}, & \text{if no pair}, \\
(R_{pc} \times 4) + S_{pc} + 1000, & \text{if there is a pair}.
\end{cases}
\end{equation} 
where $R_{pc}$ is the rank of the private card and $S_{pc}$ is the suit of the private card. 
%This allows us to generate a unique score for every possible hand using our rank and suit ranking described earlier.  
%
%The is\_bluff\_attempt function then determines whether a particular action qualifies as an attempted bluff. Specifically, 
If an agent performs a raise action while holding a hand with a strength score of 32 or lower (less than 10s and no pairs), then the action is classified as an attempted bluff. 
If the opponent folds immediately to the agents bluff attempt, then we classify that as a successful bluff. 
This rule-based detection is used to  count bluff attempts during evaluation games.
%, which enables a comparative analysis between the agents.

%\subsection{Statistics-based Bluff Detection}\label{Methodology Bluff Statistics}

In addition to the threshold-based bluff detection we also define a statistics-based bluff detection relying on belief distributions and expected values. This definition can be extended to all types of poker. 

Let $D$ be the deck, $H\subseteq D$ the set of private hands for a player, and let $pc$ denote the public context (e.g., public card, betting round, position, and other publicly observed features). Let $A$ be the action set (check, call, bet, raise, fold). 
We write $s(h)\in \mathbb{R}$ as a hand–strength function. It assigns a real number to any hand $h$ and measures how strong that hand is. Two ways of defining $s(h)$ are: 
\begin{itemize}
    \item Showdown equity: The probability of the player winning at showdown against all possible opponent hands.
    \item Normalized strength index: Assign fixed numbers to hand types, ranked from weakest to strongest. This is how we define $s(h)$ in this paper.
\end{itemize}
Let $u(h,a)\in\mathbb{R}$ denote the expected utility (EV) of taking action $a$ with hand $h$ at $pc$. We write $a_{\text{passive}}\in\{check,call\}$ for a non‑aggressive alternative available at the same information state. 
From the perspective of Player~$i$ observing Player~$j$ take action $a$ at context $pc$, let
\[
\mu(h' \mid a, pc)
\]
be Player~$i$'s belief distribution over Player~$j$'s possible private hands $h'\in H$ after observing $a$ and $pc$. We use the shorthand $h' \sim \mu(\cdot)$ to denote that $h'$ is a sample from the specified belief distribution.
Thus, a player holding $h$ is attempting to bluff with action $a$ at context $pc$ if the action strategically misrepresents strength and is EV‑preferred:
\begin{equation}
s(h) \;<\; \mathbb{E}_{h' \sim \mu(h' \mid a, pc)}\!\big[s(h')\big]
\quad \text{and} \quad
u(h,a) \;>\; u\!\big(h, a_{\text{passive}}\big)\
\end{equation}
The first clause formalizes misrepresentation: the action $a$ is more typically associated (in the observers beliefs) with stronger hands than the bluffer actually holds. The second clause requires that the bluff is rational (higher EV than the passive alternative).
Hand strength $s(h)$ is calculated as it is in the threshold-based bluff detection using the same equation.
%Our implementation in this paper slightly diverges from the formal definition above due to mainly data sparsity and computational limits. With 52 possible hands across numerous contexts, many hand-action-context combinations would have insufficient observations for reliable probability estimates. Thus, our implementation approximates the formal definition.

%For each decision context $pc$ (public card, betting round, position), and action $a$, we aggregate the observed strengths of hands $s(h)$ that took action $a$. Hand strength $s(h)$ is a deterministic index over rank and suit, with an indicator for whether the private card pairs with the public card. This yields a frequency-based "belief" summary $\hat{\mu}(h' | a, pc)$ consisting of counts and moment statistics over pair and non-pair hands (means, standard deviations, and pair frequency). We require at least five samples to use a context. We also estimate action values empirically as \[
%\hat{u}(h,a) \;=\; \mathbb{E}\!\left[ \text{payoff} \;\middle|\; pc,\, h,\, a \right]
%\] from the same logs, and we require at least three samples per $(pc, h, a)$. For context $pc = \text{(public card, betting round, position)}$, we use the public card information (if available), which betting round it is (0 for pre-flop, 1 for post-flop), and the position (simplified to player id, 0 for DQN and 1 for CFR). 
%

Note that our implementation approximates this formal definition, due to the computational intensity of producing the full belief distribution. We compute the mean and standard deviation of the strengths of hands $s(h)$ that took action $a$ at a decision context. Pair and non-pair cases are kept separate. For pairs, as the misrepresentation clause we use $\text{mean} - \sigma \times \text{threshold}$ where $\sigma$ is the standard deviation and threshold is 0.5 but it can be modified to fit specific criteria, to give a number which if the actual hand is below this number, then it is classified as misrepresentation. For non-pairs if >70\% of raises are with pairs, any non-pair raise satisfies as misrepresentation. Otherwise, we use the same method as in the pair case, but we use numbers from non-pair cases. We use payoffs as EV. 
Further details of our methods are in \cite{zaciragic2025} and our implementation codebase can be found at \cite{Zaciragic2025BluffingCodebase}.

%The misrepresentation condition is evaluated differently for when a player has pairs versus non-pairs. For non-pairs, if the empirical raise frequency in context $pc$ shows $>70\%$ pairs, then any non-pair raise is flagged as misrepresentation. If there is $<70\%$ raise frequency for pairs in a specific context, then we compare against the non-pair distribution using a threshold of $\text{mean} - \sigma \times \text{threshold}$ where $\sigma$ is the standard deviation and we use a threshold of 0.5 but it can be modified to fit specific criteria. If the player has pairs then we compare against the pair distribution using the same threshold. This is implemented in a two-pass pipeline, the first pass builds $\hat{\mu}$ and $\hat{u}$ for all $(pc, a)$, and the second pass classifies each raise and tallies attempt/success rates and opponent reactions.

%Expected utilities $u(h,a)$ are estimated by tracking the final payoffs for each $(h,a,pc)$ combination across games. The EV condition requires at least 3 samples for reliability and checks whether $u(h,\text{raise}) > u(h,\text{call})$. 

%This parametric approximation preserves the core insight that bluffing involves misrepresenting hand strength relative to empirical expectations, making it a practical and theoretically sound approach for large-scale strategic analysis. 

\section{Results}\label{Results}

%Only make inferences after you have presented the objective results, seperate with a sntadard sentence such as "Based on these results we can conclude that"

%\subsection{Training}\label{Results Training}

In the training phase, both the CFR and DQN agents were trained simultaneously against each other on  100 000 games of Leduc Hold'em with the aim of allowing both agents to adapt to eachother.

\begin{figure}[h]
    \centering
    \begin{subfigure}[b]{0.48\linewidth}
        \includegraphics[width=\linewidth]{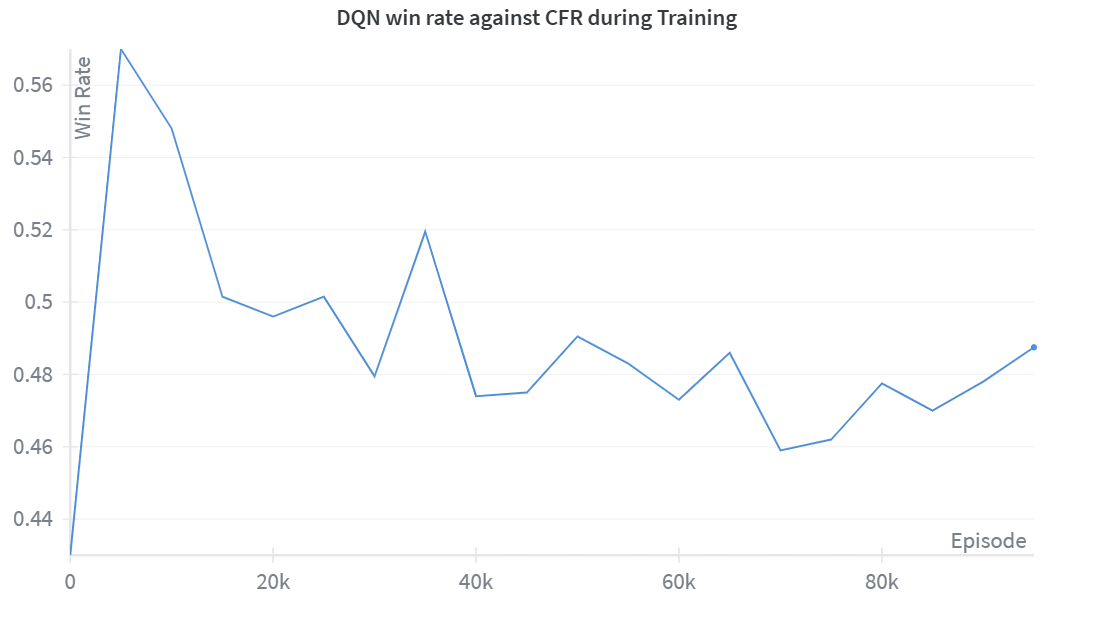}
        \caption{DQN win rate vs. CFR}
        \label{fig:DQN_vs_CFR_simultaneous}
    \end{subfigure}
    \hfill
    \begin{subfigure}[b]{0.48\linewidth}
        \includegraphics[width=\linewidth]{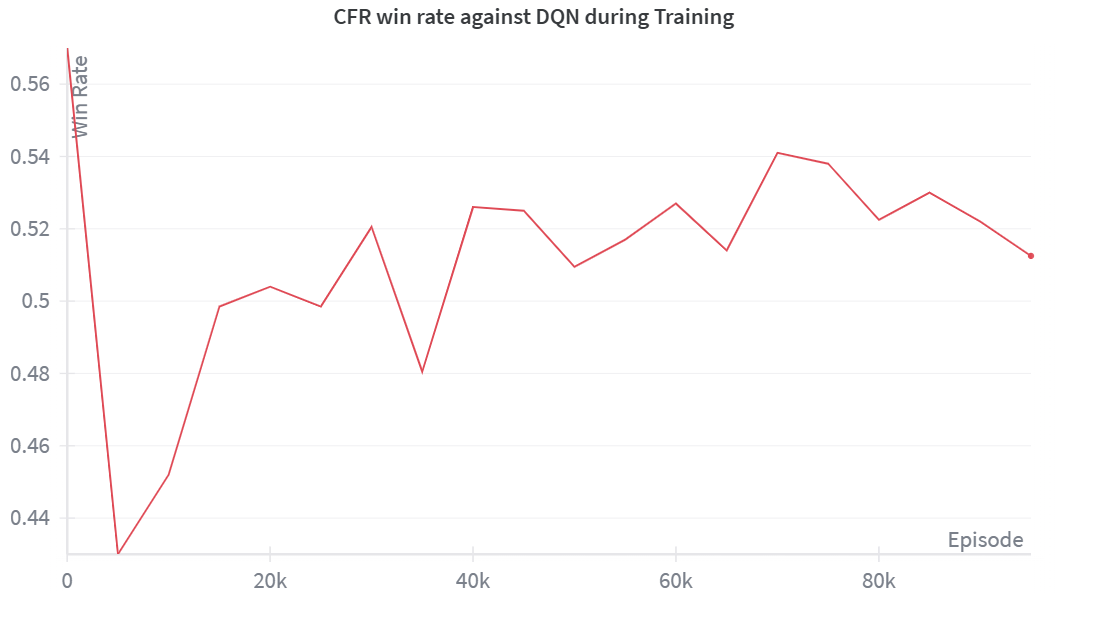}
        \caption{CFR win rate vs. DQN}
        \label{fig:CFR_vs_DQN_simultaneous}
    \end{subfigure}
    \caption{Win rate dynamics during simultaneous training of DQN and CFR agents in Leduc Hold'em. (a) DQN initially gains an edge, but its win rate declines as CFR adapts. (b) CFR steadily improves, reaching a win rate of 40–50\%.}
    \label{fig:simultaneous_training_winrates}
\end{figure}

Figure \ref{fig:simultaneous_training_winrates} illustrates the win-rate dynamics between the DQN and CFR agents during training. From Figure \ref{fig:DQN_vs_CFR_simultaneous} we can see that initially the DQN agent achieves a slight advantage, with its win rate peaking above 55\%. However, this edge quickly diminishes as the CFR agent adapts, which leads to a decline in DQNs win rate which stabilizes in the 46\%-49\% range for most of the training process. Figure \ref{fig:CFR_vs_DQN_simultaneous} shows the complementary trend, that is, CFR immediately drops from a 56\% win rate down to 44\%, but it steadily improves after this and maintain a persistent advantage with win rates ranging between 50\%-54\% during most of the training period. 

This observed difference is consistent with the theoretical properties of the two algorithms. Namely, CFR is explicitly designed for imperfect-information games and leverages regret minimization to converge towards equilibrium strategies under the right conditions. %Specifically, those conditions are two-player, zero-sum games with perfect recall, which Leduc Hold'em satisfies. 
Practical convergence to a Nash equilibrium typically requires millions of iterations, and the training budget in this paper was insufficient to guarantee equilibrium play. This can be also seen from the volatile win rate graphs which show that true convergence has not yet been reached. 

The graphs also indicate partial progress towards less-exploitable play as CFRs average strategy becomes harder to exploit than DQNs function approximation policy. Nevertheless, its regret-minimization updates systematically push strategies closer to equilibrium and eliminate highly exploitable behaviors. By contrast, DQN is optimizing a value function from sampled play against a non-stationary opponent as CFRs policy changes every episode. Off-policy TD methods such as DQN  assume a stationary target, however, here that assumption is violated, which amplifies function-approximation noise and over/under estimation. With $\epsilon$-greedy exploration and finite play, DQN probably undersamples rare but crucial situations, so its early strategy gets counter-adapted as CFRs regret updates rebalance action probabilities towards more profitable actions faster.

\subsection{Do CFR and DQN bluff?}\label{Experiments RQ2}

\begin{figure}[h]
  \centering

  % Top row
  \begin{subfigure}[t]{0.48\textwidth}
    \centering
    \includegraphics[width=\linewidth]{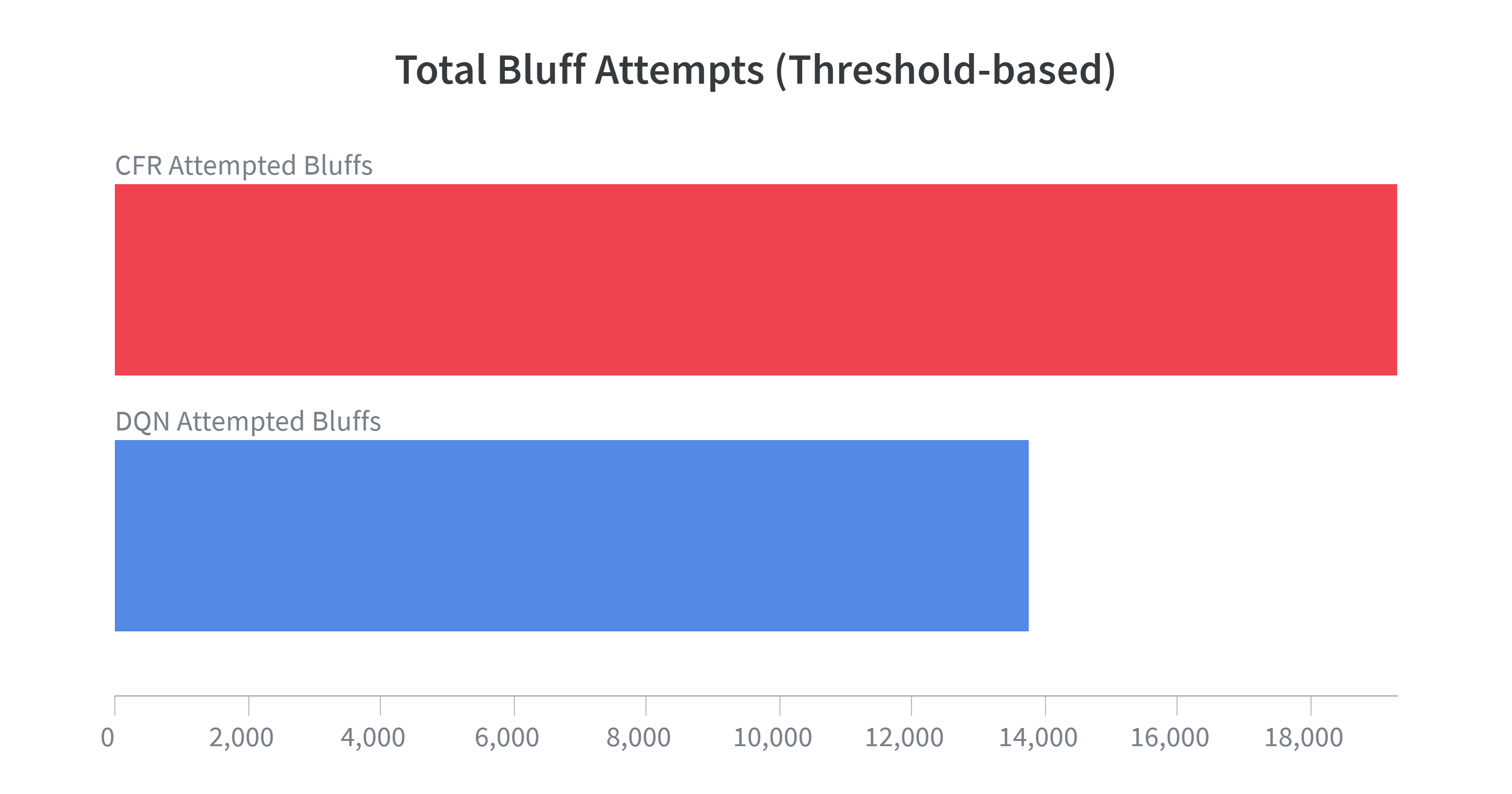}
    \caption{Total number of bluffs attempted by both DQN and CFR (using the threshold-based detector).}
    \label{fig:tl}
  \end{subfigure}
  \hfill
  \begin{subfigure}[t]{0.48\textwidth}
    \centering
    \includegraphics[width=\linewidth]{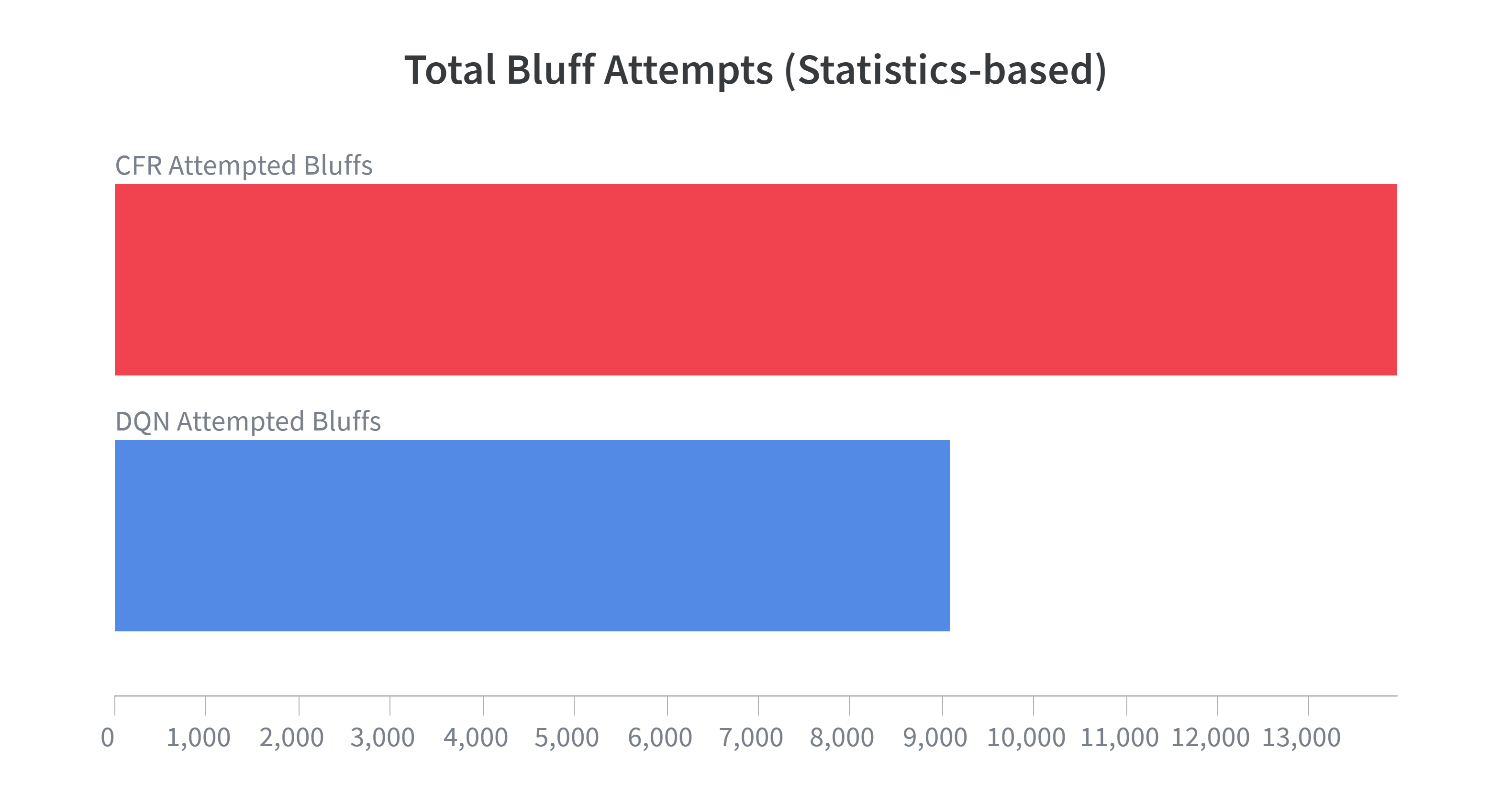}
    \caption{Total number of bluffs attempted by both DQN and CFR (using the statistics-based detector).}
    \label{fig:tr}
  \end{subfigure}

  \medskip

  % Bottom row
  \begin{subfigure}[t]{0.48\textwidth}
    \centering
    \includegraphics[width=\linewidth]{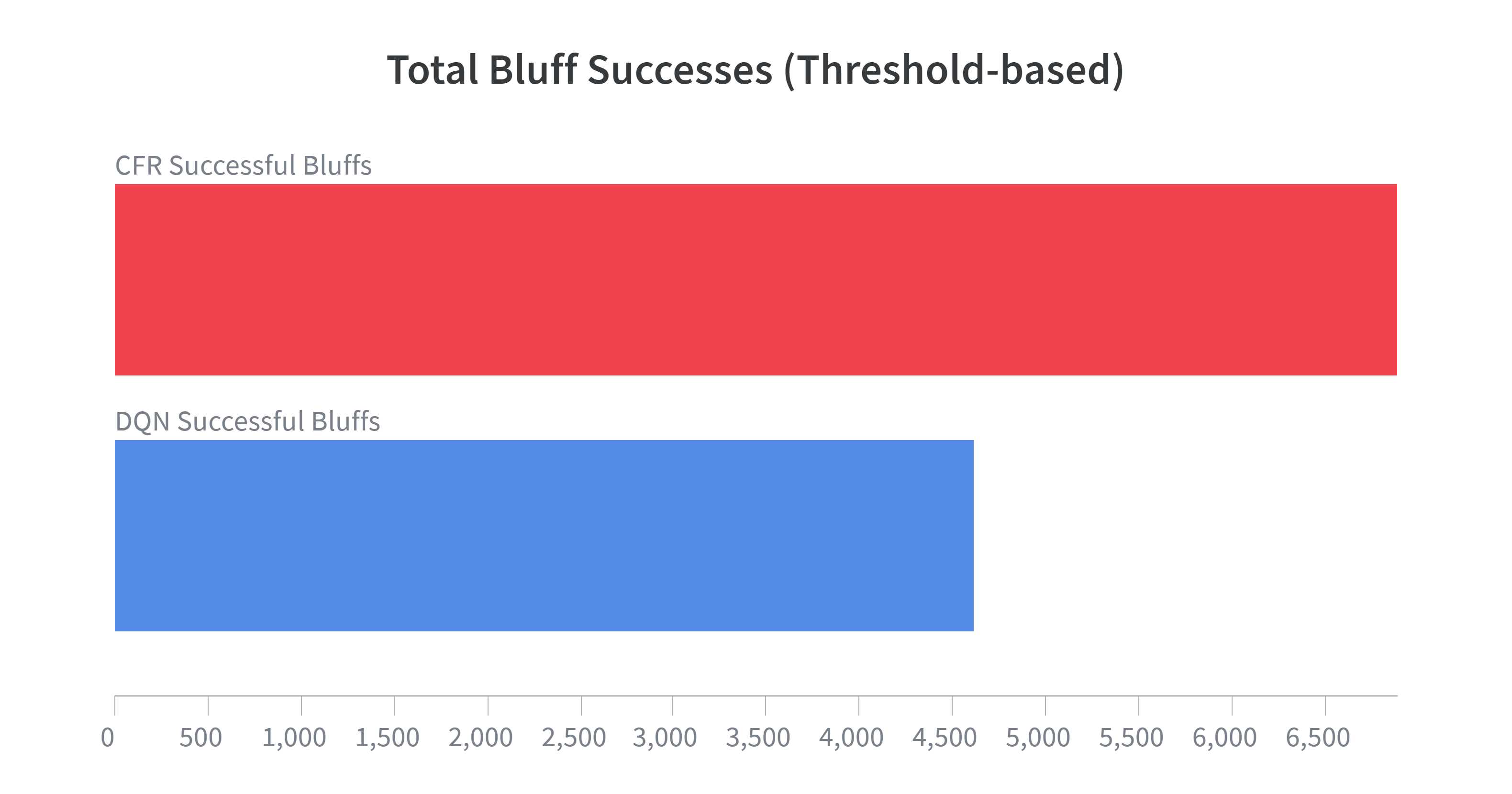}
    \caption{Total number of Successful bluffs by both DQN and CFR (using the threshold-based detector).}
    \label{fig:bl}
  \end{subfigure}
  \hfill
  \begin{subfigure}[t]{0.48\textwidth}
    \centering
    \includegraphics[width=\linewidth]{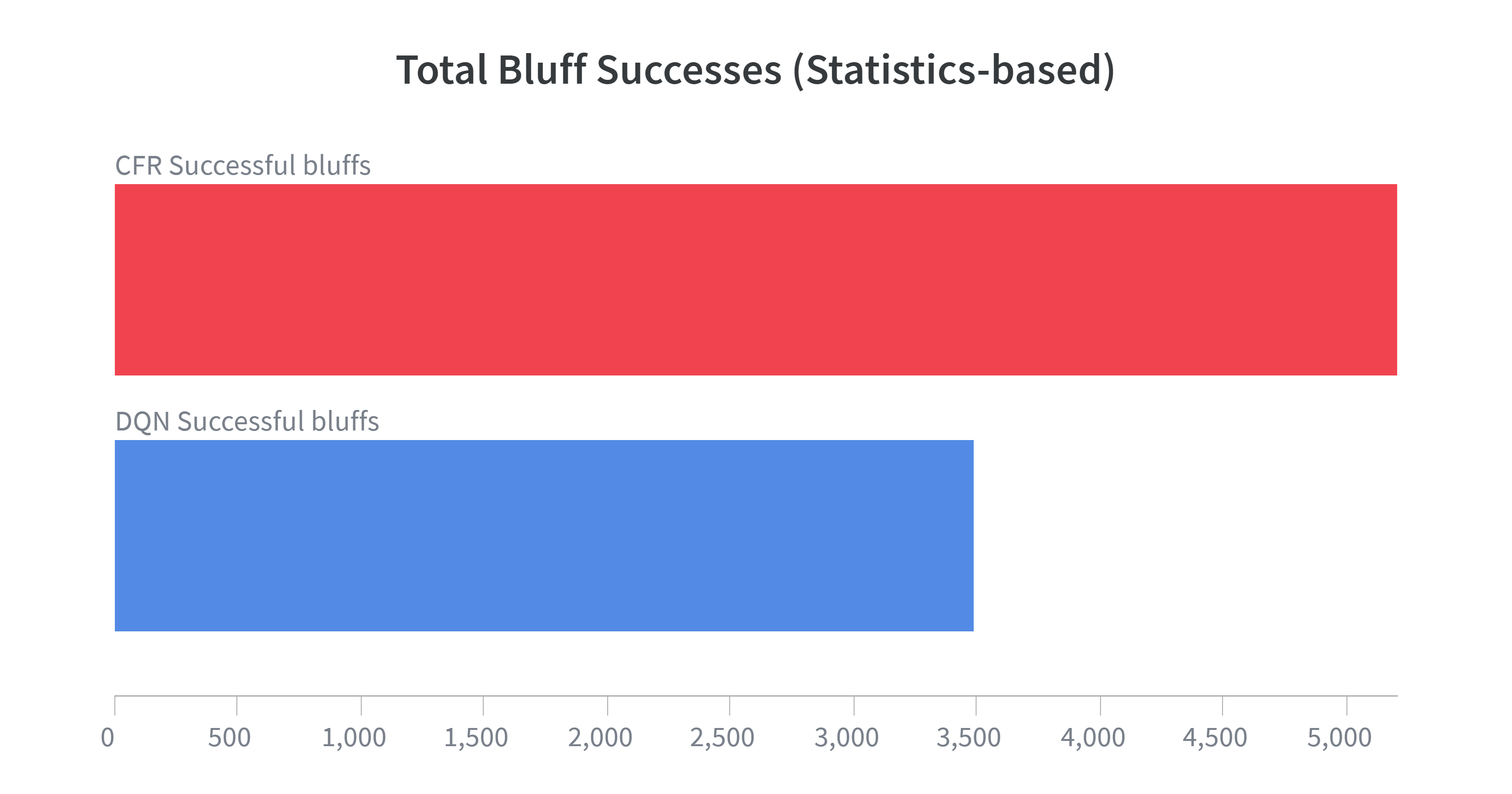}
    \caption{Total number of Successful bluffs by both DQN and CFR (using the statistics-based detector).}
    \label{fig:br}
  \end{subfigure}

  \caption{Bluff attempt and success frequency. Panels (a)–(b) report total bluff attempts under our two bluff detection methods. We can see that in both panels, (a) and (b), CFR attempts more bluffs than DQN. Panels (c)–(d) report the number of bluffs classified as successful under each detector. We can see that CFR has more succesful bluffs than DQN in absolute numbers, however, when taking into account panels (a) and (b) the success rates are very similar.}
  \label{fig: Bluffing Attempts and Successes}
\end{figure}

From Figure \ref{fig: Bluffing Attempts and Successes} we can see that both agents have a large number of both attempted and successful bluffs using two independent detectors. This shows us that indeed both DQN and CFR engage in bluffing. The statistics-based detection yields fewer counts than the simpler threshold heuristic, which is to be expected considering that the statistics-based one has a stricter definition of bluffing. 

A possible explanation of the differences in bluffing behaviour between the two algorithms is that these differences can be attributed to the paradigms that each agent originates from. CFR attempts to bluff  more than DQN because CFR is equilibrium-driven and it must sometimes bluff to remain unpredictable. %When an agent does not bluff at all the agent is predictable and can be exploited. When an agent bluffs all the time, again, the agent is predictable and can be exploited. Thus, there needs to be an optimal amount and interval of bluffing in order to remain unexploitable. 
On the other hand, DQN only bluffs when the estimated Q-value says it is profitable. Since it is not taught about bluffing explicitly, it tends to be more conservative with its attempts. Thus, it under-bluffs compared to CFR. 

Interestingly, the  overall success rates are similar despite the difference in the number of attempts. The bluff success rate for CFR is 36\% by the threshold-based detector and 37\% by the statistical-based detector, and for DQN it is 34\% by the threshold-based detector and 39\% by the statistical-based detector. The success rate depends on the willingness of the opponent to fold. Both CFR and DQN learn their folding frequencies during training. CFR learns them as a part of minimizing regret and if folding too often would allow bluffs to exploit it, CFR would reduce folding in those states. If calling too often would lose against value bets, then CFR would increase folding. On the other hand, DQN learns to fold implicitly through its Q-values. If calling has a negative expected value and folding has a higher expected value, the Q-network will assign a higher value to folding. The similar  bluff success rate shows that both DQN and CFR on the opponent side have developed a comparable level of strategic competence. This is similar to human poker dynamics where players of similar skill levels will tend to achieve similar bluff success rates because their ability to detect and respond to deceptive play is comparably refined. 

%Additionally, due to the agents achieving a similar success rate with different attempt rates, we can say that success rates and attempt rates are independent of each other, that is, the success rate does not depend on the amount of times an agent has attempted a bluff. However, we also think that both agents have reached a "equilibrium" bluffing attempt frequency where if the agent were to attempt bluffs much more or much less, its plays would become more predictable, and the opponent could adjust to it by calling or folding more frequently, and thereby changing the success rate. 

%Figure \ref{fig: Bluff Attempts vs Successes by Rank} shows the details of the bluffing attempts and successes of both agents. 

\begin{figure}[!ht]
  \centering

  % Top row
  \begin{subfigure}[t]{0.48\textwidth}
    \centering
    \includegraphics[width=\linewidth]{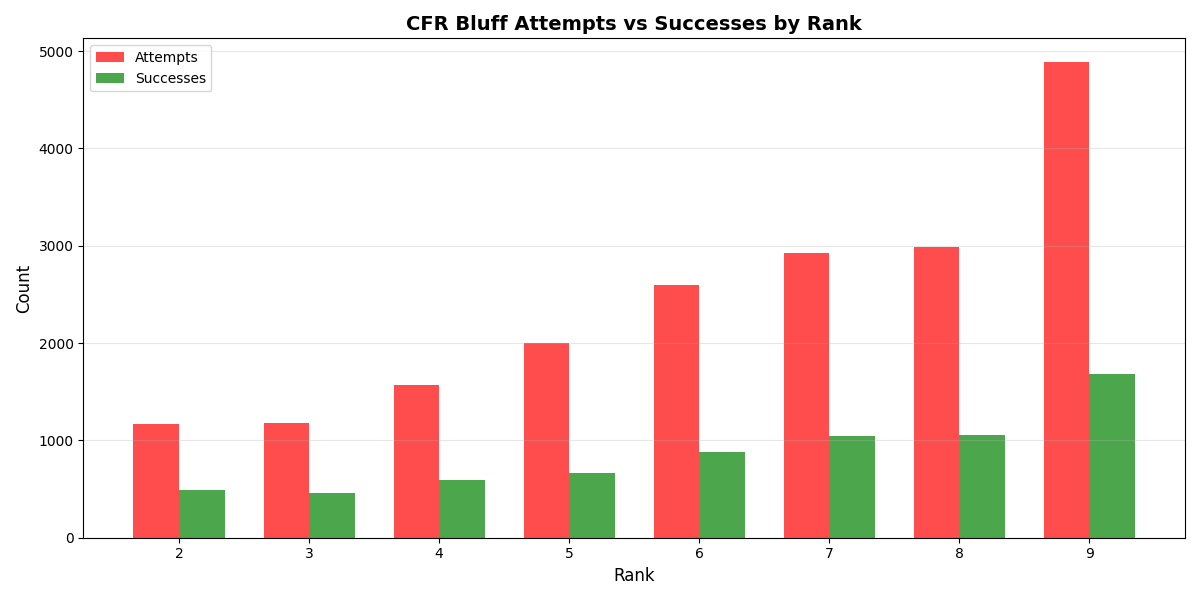}
    \caption{CFR Number of Bluff Attempts and Bluff Successes by Card Rank (using the threshold-based detector).}
    \label{fig: CFR Threshold Detector Bluff Attempts vs Successes by Rank}
  \end{subfigure}
  \hfill
  \begin{subfigure}[t]{0.48\textwidth}
    \centering
    \includegraphics[width=\linewidth]{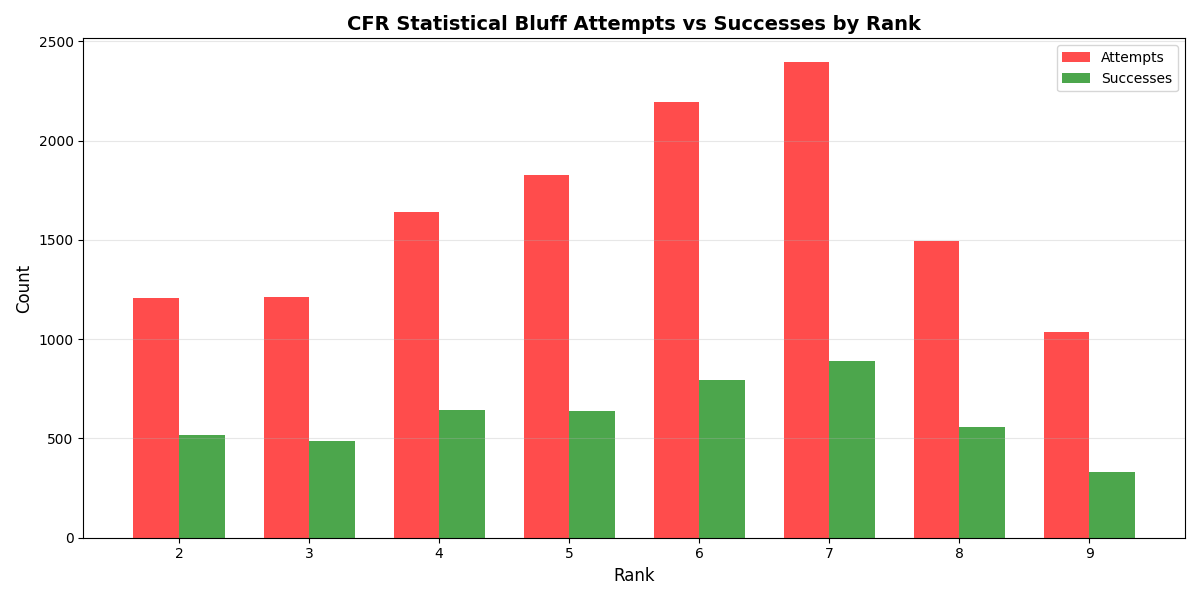}
    \caption{CFR Number of Bluff Attempts and Bluff Successes by Card Rank (using the statistics-based detector).}
    \label{fig: CFR Statistics Detector Bluff Attempts vs Successes by Rank}
  \end{subfigure}

  \medskip

  % Bottom row
  \begin{subfigure}[t]{0.48\textwidth}
    \centering
    \includegraphics[width=\linewidth]{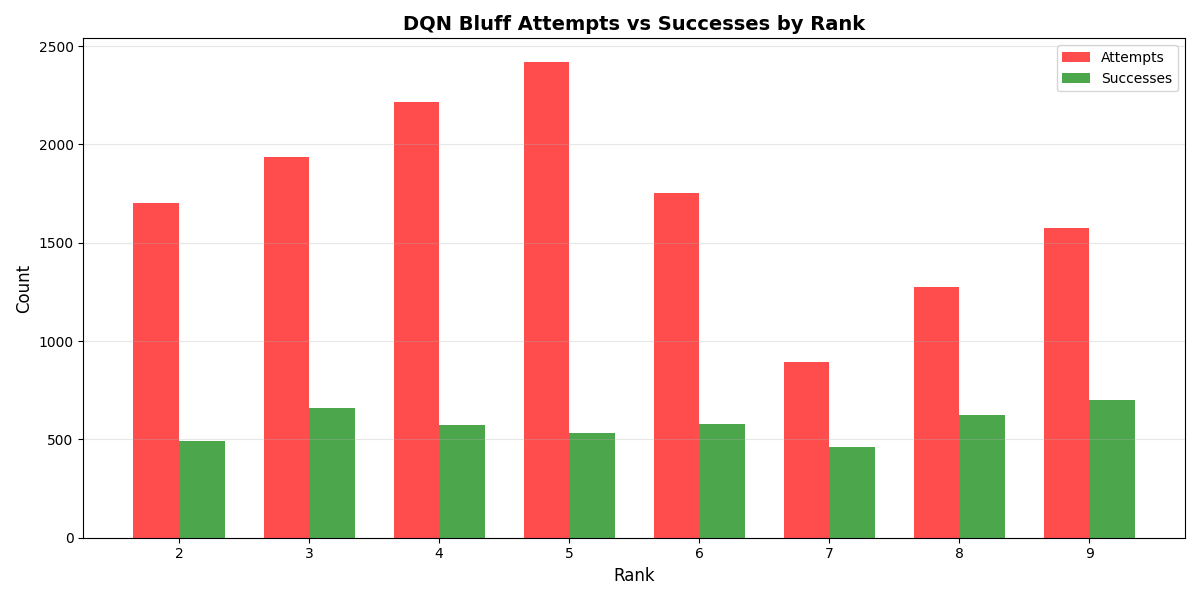}
    \caption{DQN Number of Bluff Attempts and Bluff Successes by Card Rank (using the threshold-based detector).}
    \label{fig: DQN Threshold Detector Bluff Attempts vs Successes by Rank}
  \end{subfigure}
  \hfill
  \begin{subfigure}[t]{0.48\textwidth}
    \centering
    \includegraphics[width=\linewidth]{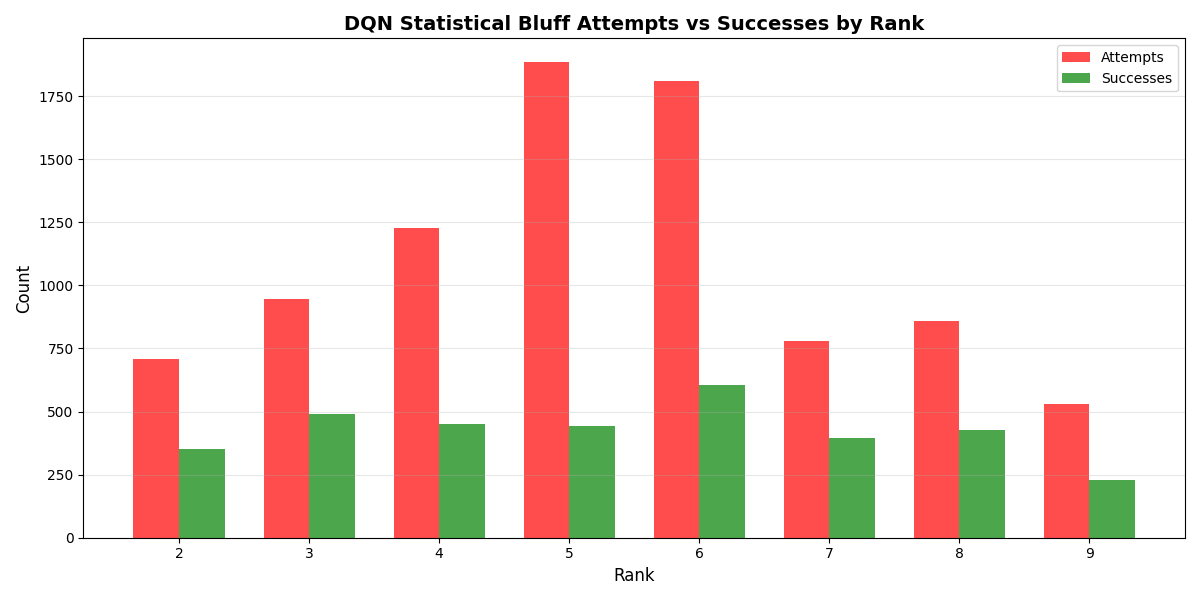}
    \caption{DQN Number of Bluff Attempts and Bluff Successes by Card Rank (using the statistics-based detector).}
    \label{fig: DQN Statistics Detector Bluff Attempts vs Successes by Rank}
  \end{subfigure}

  \caption{Bluff attempt and success frequency by card rank. Panels (a)-(b) show the comparison between the number of attempted vs. the number of successful bluffs by CFR under our two detectors. Panel (a) shows that bluff attempts increase with rank, peaking at ranks 7–9 and successes scale proportionally. Panel (b) shows a similar trend, but the attempts peak at ranks 5-7 and then fall off, with successes following a similar pattern. Panels (c)-(d) show the comparison between the number of attempted vs. the number of successful bluffs by DQN under our two detectors. From (c), we can see how attempts scale with rank peaking at ranks 4-5 and then fall off before going up in numbers again. Successes hover around the same number across ranks. From (d), we can see that attempts again scale with rank, peaking at ranks 5-6 and then fall off. The successes hover around the same number but this time with more variation between ranks.}
  \label{fig: Bluff Attempts vs Successes by Rank}
\end{figure}

Figure \ref{fig: CFR Threshold Detector Bluff Attempts vs Successes by Rank} shows that CFRs bluffing attempts increase in size from 2 (the lowest rank) to 9 (the highest rank considered to be a bluff, this is where the cutoff is). The successes roughly scale in the same way, although, with smaller size. This shows that CFR does not only bluff when having a very weak hand, but attempts to bluff also with mid to high-rank cards. CFR's strategy is systematic and integrated into its overall strategy. Most contiguous ranks have similar attempt sizes, which indicates that CFR is preventing to be exploited by distributing its bluffing attempts across many cards. The conversion rate is also roughly the same across most ranks, which indicates that opponents cannot easily exploit CFR's bluffs.
%because they are mixed in across ranks which makes it hard to predict when it is a bluff and when its not. 
The results obtained from the statistics-based detector shown in figure \ref{fig: CFR Statistics Detector Bluff Attempts vs Successes by Rank}  follow the same trend, but with smaller absolute sizes of attempts and successes due to the stricter classification. 
%The distribution is also skewed to the left. 
Both graphs show that CFR bluffs as humans do in real life poker, that is, they choose to bluff with hands that are weak enough to probably lose at showdown but still have some chances of winning as a high card.

On the other hand, Figure \ref{fig: DQN Threshold Detector Bluff Attempts vs Successes by Rank} reveals that the DQN agent bluffs most often in the mid-rank region (3-6). These are the hands that are quite weak and very likely to lose at showdown, but are used by the DQN agent to stay in the game through bluffing with them. The success rate is about the same across ranks, but 
%the success rate is also a property of the opponent as it 
also depends on whether the opponent has learned to associate certain patterns with certain hands and therefore to certain actions such as folding. DQN does not have any built-in concepts such as deception or bluffing. It bluffs only because Q-learning  rewards certain aggressive actions that occasionally yield higher expected return than folding outright. %DQN might have discovered that sometimes being aggressive with weak hands still earns positive expected value, especially when CFR folds enough. 
The possible reason why the bluffing clusters in the mid-rank zone is because the risk-reward trade-off has showed up the most in these zones in its experience. Additionally, training against CFR also shapes its bluffing style. CFR is trying to minimize its own exploitability which prevents DQN from becoming an all-in-bluffer because DQN would lose badly. Instead, it pushes DQN to find a more selective group of hands where bluffing still works against CFR. Figure \ref{fig: DQN Statistics Detector Bluff Attempts vs Successes by Rank} shows a very similar trend, the biggest difference being the smaller quantity of games due to stricter classification. 

%Threshold-based reactions
\begin{figure}[!t]
  \centering

  % First row
  \begin{subfigure}[t]{0.32\textwidth}
    \centering
    \includegraphics[width=\linewidth]{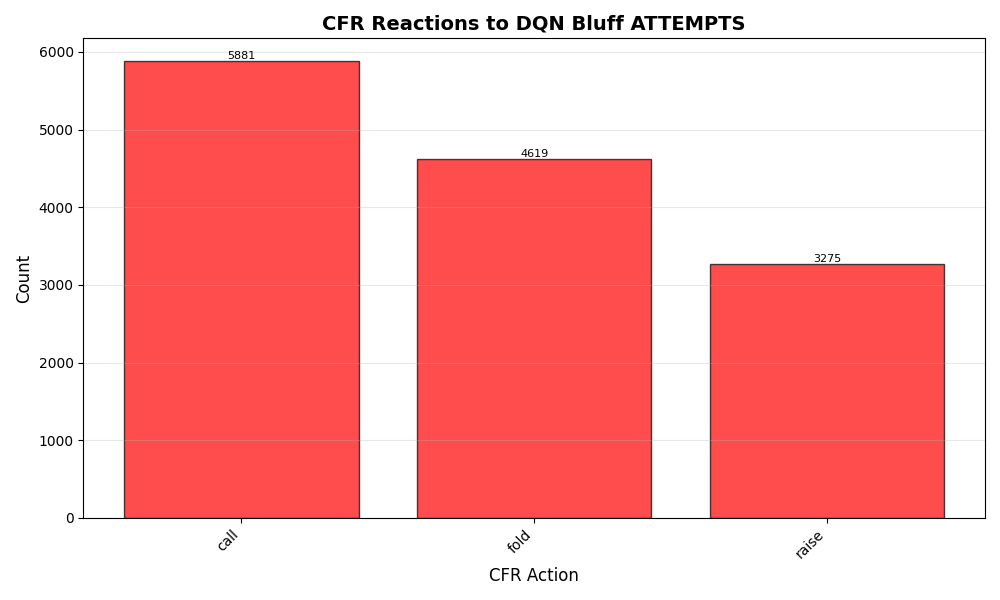}
    \caption{CFR’s overall reactions to DQN bluffs, aggregated across all phases (using the threshold-based detector).}
    \label{fig: CFR Reactions Threshold-based}
  \end{subfigure}
  \hfill
  \begin{subfigure}[t]{0.32\textwidth}
    \centering
    \includegraphics[width=\linewidth]{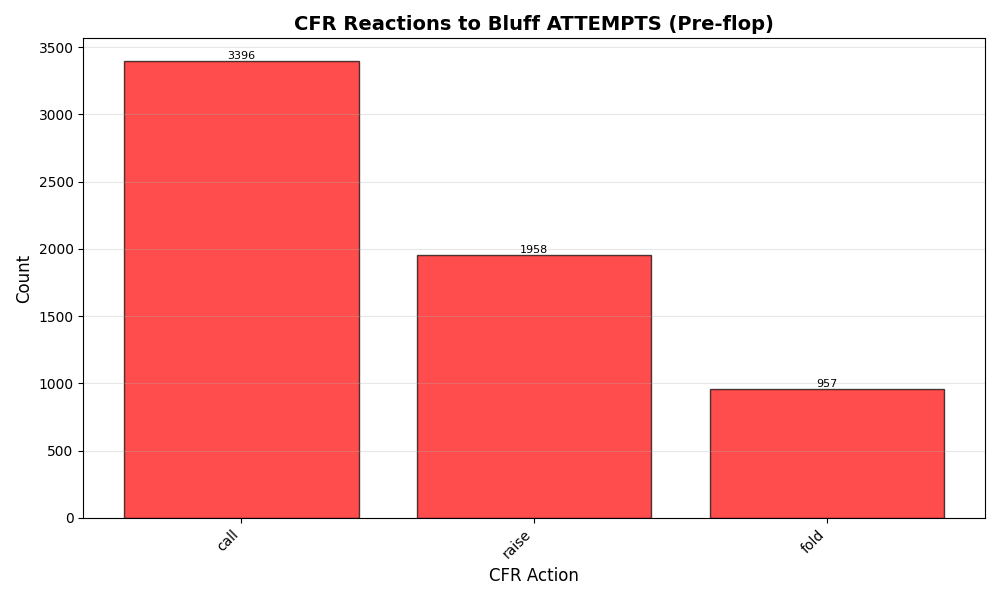}
    \caption{CFR’s reactions to DQN bluffs during the pre-flop stage (using the threshold-based detector).}
    \label{fig: CFR Reactions PreFlop Threshold-based}
  \end{subfigure}
  \hfill
  \begin{subfigure}[t]{0.32\textwidth}
    \centering
    \includegraphics[width=\linewidth]{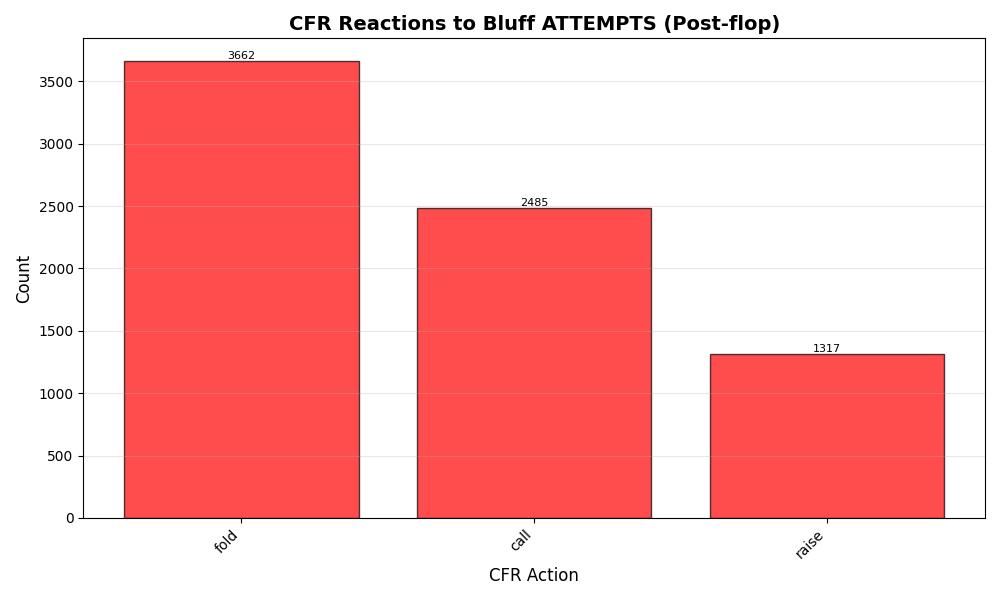}
    \caption{CFR’s reactions to DQN bluffs during the post-flop stage (using the threshold-based detector).}
    \label{fig: CFR Reactions PostFlop Threshold-based}
  \end{subfigure}

  \medskip

  % Second row
  \begin{subfigure}[t]{0.32\textwidth}
    \centering
    \includegraphics[width=\linewidth]{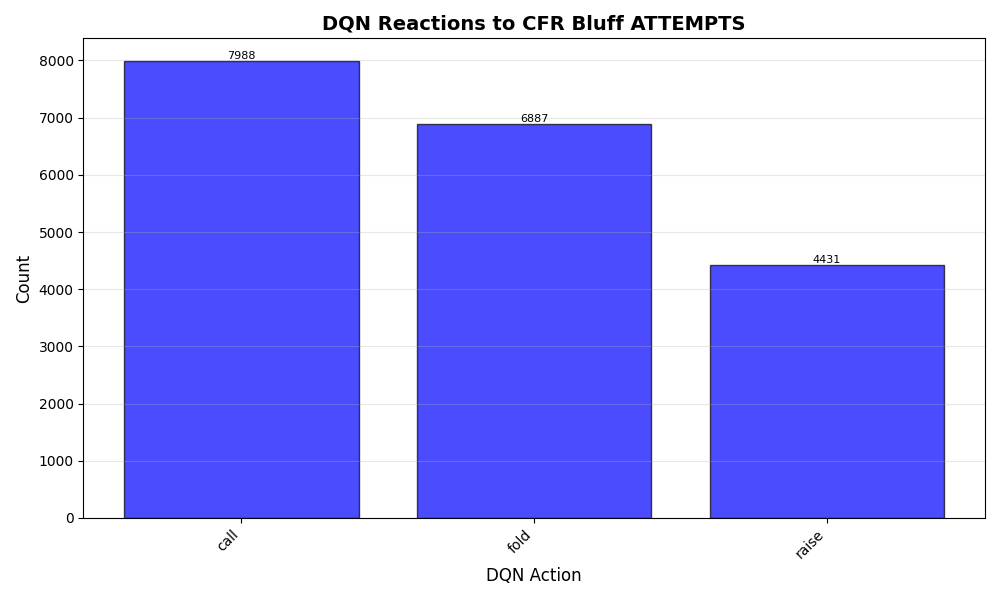}
    \caption{DQN’s overall reactions to CFR bluffs, aggregated across all phases (using the threshold-based detector).}
    \label{fig: DQN Reactions Threshold-based}
  \end{subfigure}
  \hfill
  \begin{subfigure}[t]{0.32\textwidth}
    \centering
    \includegraphics[width=\linewidth]{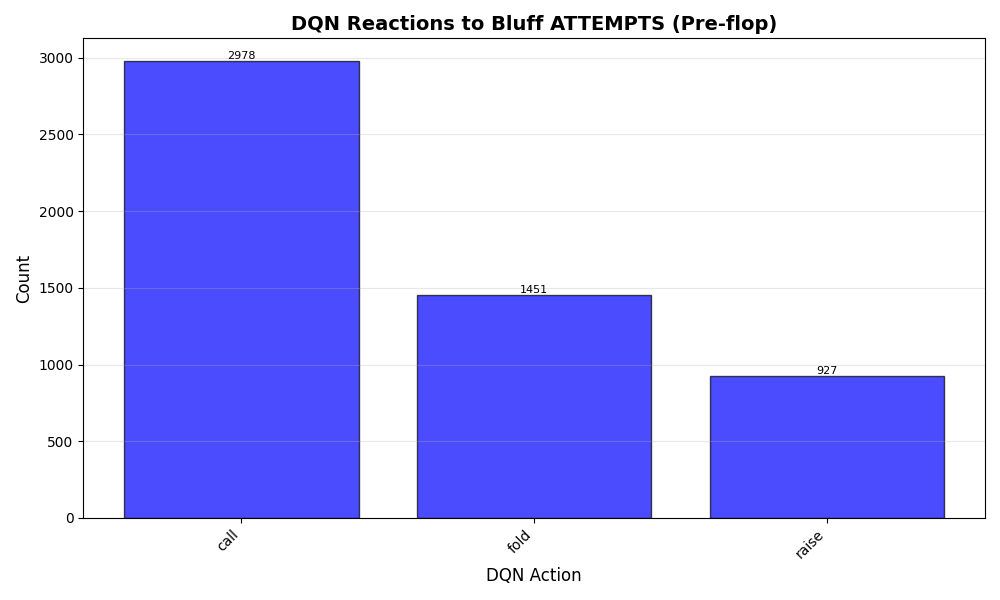}
    \caption{DQN’s reactions to CFR bluffs during the pre-flop stage (using the threshold-based detector).}
    \label{fig: DQN Reactions PreFlop Threshold-based}
  \end{subfigure}
  \hfill
  \begin{subfigure}[t]{0.32\textwidth}
    \centering
    \includegraphics[width=\linewidth]{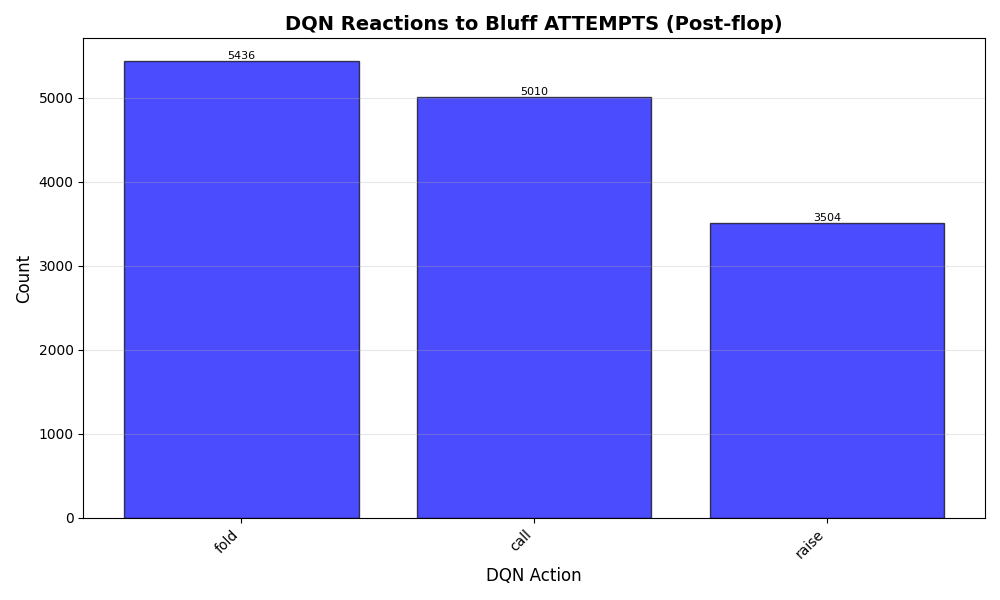}
    \caption{DQN’s reactions to CFR bluffs during the post-flop stage (using the threshold-based detector).}
    \label{fig: DQN Reactions PostFlop Threshold-based}
  \end{subfigure}

  \caption{Opponent reactions to bluffing under the threshold-based detector. Panel (a) highlights how frequently CFR chooses to fold, call, or raise in response to DQN's bluff attempts. Panels (b)-(c) show the same just across the two different phases of the game. CFR prefers to call overall and in the pre-flop stage, while in the post-flop stage it prefers to fold. Panel (d) reveals how often DQN chooses to fold, call, or raise in response to CFR's bluff attempts. Panels (e)-(f) show the same just across the two different stages of the game. DQN prefers to call overall and in the pre-flop stage, while preferring to fold in the post-flop stage. Both agents react similarly.}
  \label{fig: Opponent Reactions Threshold-based}
\end{figure}

%Statistics-based Reactions
\begin{figure}[!t]
  \centering

  % First row
  \begin{subfigure}[t]{0.32\textwidth}
    \centering
    \includegraphics[width=\linewidth]{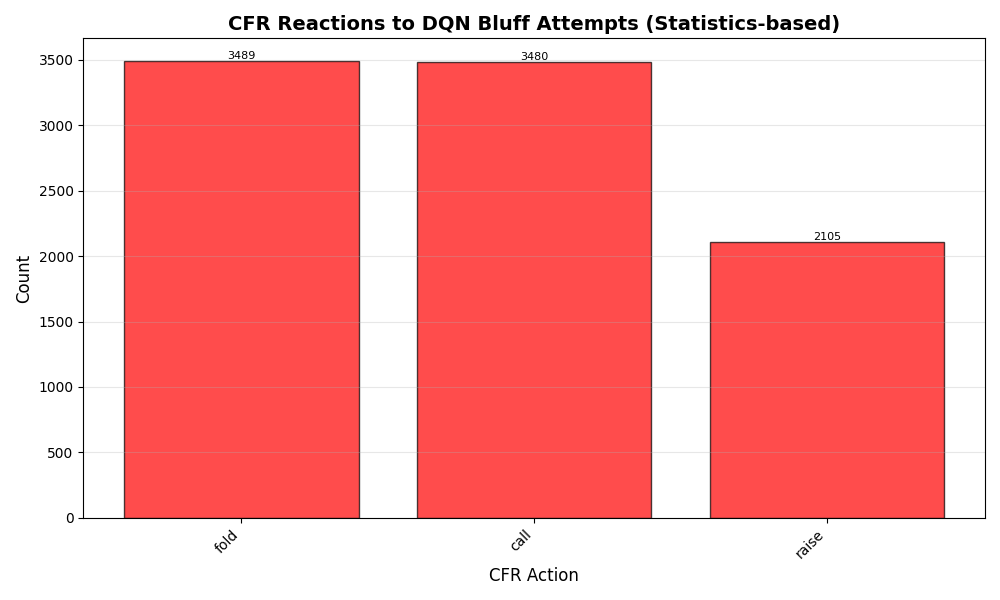}
    \caption{CFR’s overall reactions to DQN bluffs, aggregated across all phases (using the statistics-based detector).}
    \label{fig: CFR Reaction Statistics-based}
  \end{subfigure}
  \hfill
  \begin{subfigure}[t]{0.32\textwidth}
    \centering
    \includegraphics[width=\linewidth]{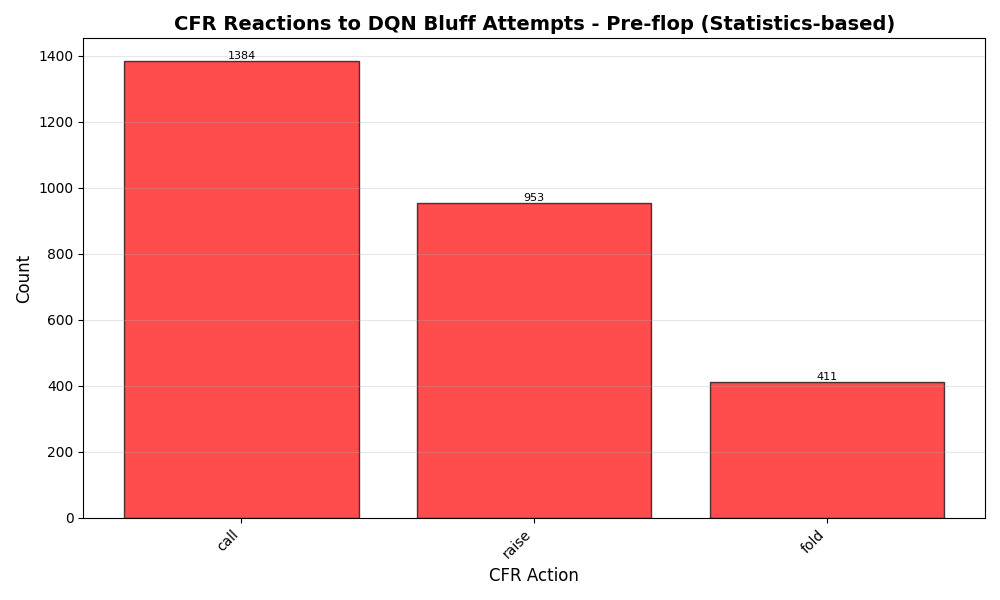}
    \caption{CFR’s reactions to DQN bluffs during the pre-flop stage (using the statistics-based detector).}
    \label{fig: CFR Reaction PreFlop Statistics-based}
  \end{subfigure}
  \hfill
  \begin{subfigure}[t]{0.32\textwidth}
    \centering
    \includegraphics[width=\linewidth]{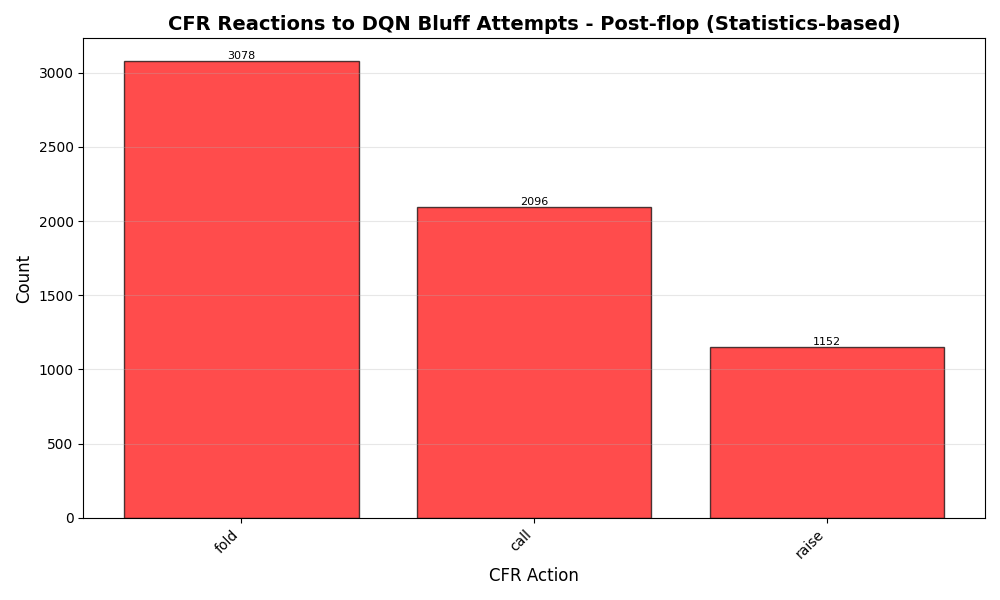}
    \caption{CFR’s reactions to DQN bluffs during the post-flop stage (using the statistics-based detector).}
    \label{fig: CFR Reaction PostFlop Statistics-based}
  \end{subfigure}

  \medskip

  % Second row
  \begin{subfigure}[t]{0.32\textwidth}
    \centering
    \includegraphics[width=\linewidth]{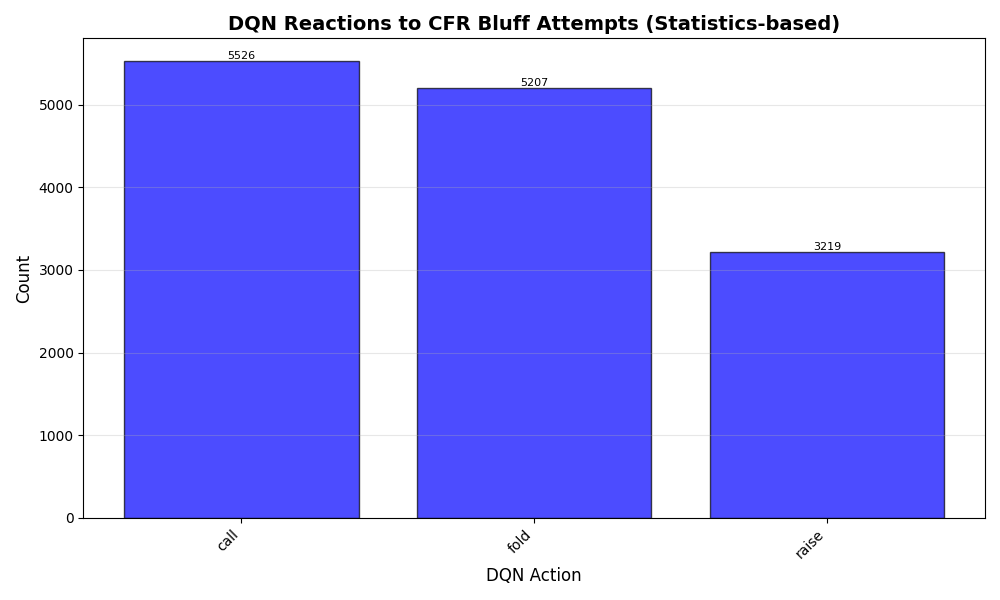}
    \caption{DQN’s overall reactions to CFR bluffs, aggregated across all phases (using the statistics-based detector).}
    \label{fig: DQN Reaction Statistics-based}
  \end{subfigure}
  \hfill
  \begin{subfigure}[t]{0.32\textwidth}
    \centering
    \includegraphics[width=\linewidth]{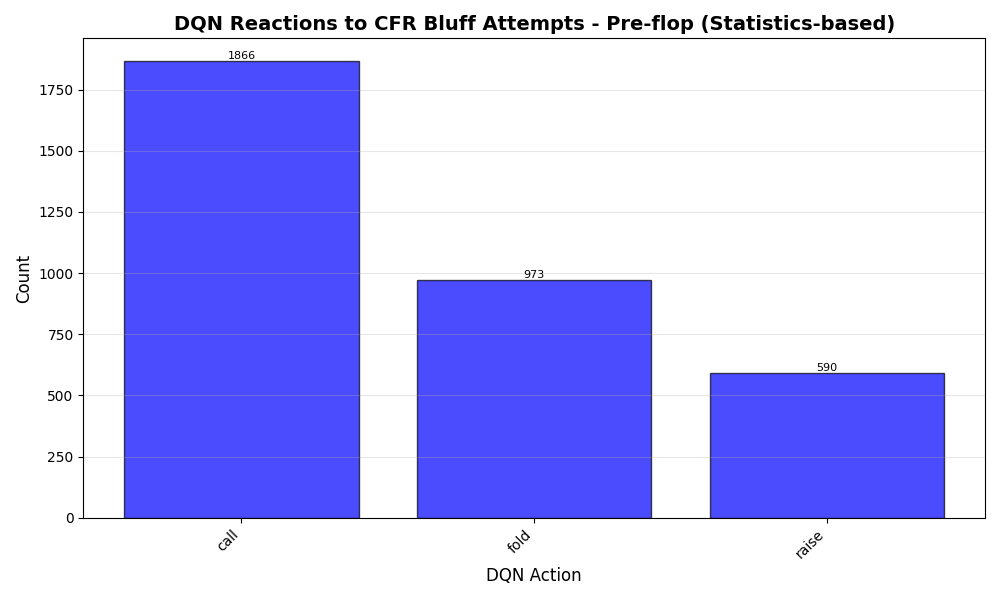}
    \caption{DQN’s reactions to CFR bluffs during the pre-flop stage (using the statistics-based detector).}
    \label{fig: DQN Reaction PreFlop Statistics-based}
  \end{subfigure}
  \hfill
  \begin{subfigure}[t]{0.32\textwidth}
    \centering
    \includegraphics[width=\linewidth]{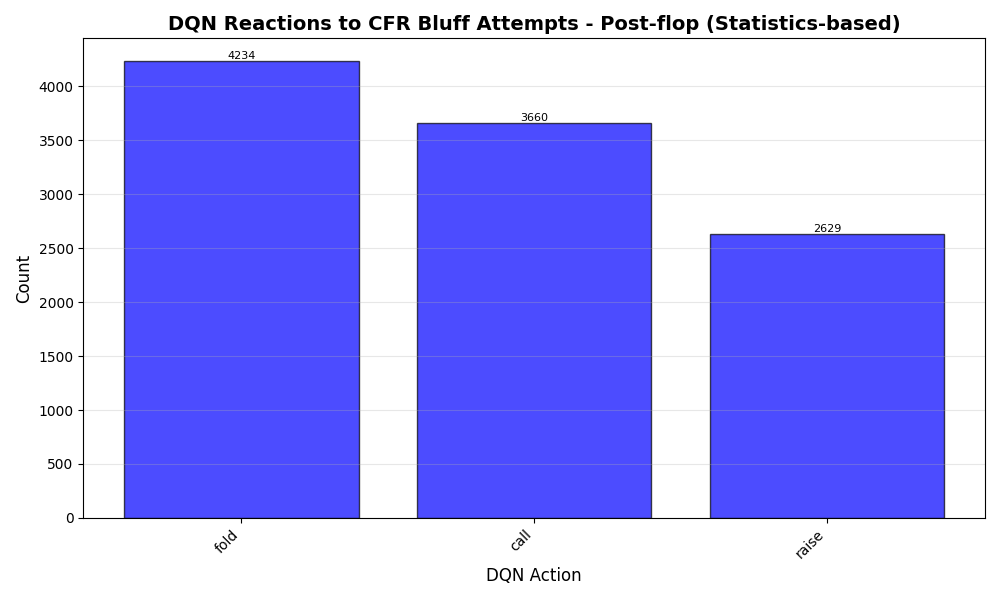}
    \caption{DQN’s reactions to CFR bluffs during the post-flop stage (using the statistics-based detector).}
    \label{fig: DQN Reaction PostFlop Statistics-based}
  \end{subfigure}

  \caption{Opponent reactions to bluffing under the statistics-based detector. Panel (a) highlights how frequently CFR chooses to fold, call, or raise in response to DQN's bluff attempts. Panels (b)-(c) show the same just across the two different phases of the game. CFR prefers to call and fold overall. Furthermore, in the pre-flop stage it prefers to call, while in the post-flop stage it prefers to fold. Panel (d) reveals how often DQN chooses to fold, call, or raise in response to CFR's bluff attempts. Panels (e)-(f) show the same just across the two different stages of the game. DQN prefers to call overall and in the pre-flop stage, while preferring to fold in the post-flop stage. Both agents react very similarly.}
  \label{fig: Opponent Reactions Statistics-based}
\end{figure}

Figure \ref{fig: CFR Reactions Threshold-based} shows that the most common action of CFR to DQNs bluffing attempts is calling, followed by folding and raising. This indicates that CFR chooses to see through the bluff rather than retreat (fold) or counterattack (reraise). This might be done in order to minimize regret by avoiding to fold too early. CFR folds when the expected risk is too high to avoid leaking chips. It might raise when it is extremely confident that the opponent is bluffing in order to make the opponent fold. To get a more nuanced understanding of the reactions, we also analyze the reactions per game phase. Figure \ref{fig: CFR Reactions PreFlop Threshold-based} reveals that before the public card is shown CFR prefers to be in the game and gather more information rather than to fold prematurely. It also likes to reraise, possibly to scare off the opponent. Lastly, it folds the least. However, when looking at the post-flop phase from figure \ref{fig: CFR Reactions PostFlop Threshold-based}, once the public card is revealed, CFR prefers to fold rather than to raise or call. This indicates that CFR becomes more conservative once more information is available, choosing to fold in unfavorable situations or when DQNs bets appear to be strong. The statistics-based detector produces similar results as can be seen in Figures \ref{fig: CFR Reaction Statistics-based}, \ref{fig: CFR Reaction PreFlop Statistics-based} and \ref{fig: CFR Reaction PostFlop Statistics-based}. The statistics-based detector is used to verify that the general patterns are still there even when a different way of detecting bluffs is used. The main difference is that there is less cases in the statistics-based detector graphs. Folding also becomes the main overall action, but since it is only by a few cases, it can be plausibly attributed to chance.

DQN follows a similar pattern as CFR, preferring to call most of the bluffing attempts, then to fold and then to raise. This can be seen from Figure \ref{fig: DQN Reactions Threshold-based} which  follows the  same pattern as CFR's reactions but with more cases. It is not surprising that DQN prefers to call the most in its response since calling is often the least punishing action because it allows the agent to gather more information about outcomes (to go and see showdowns). Folding would cut off rewards entirely (as you give up and do not see your opponents cards) and raising would introduce higher variance that could destabilize value estimation. DQN folds only when the learned Q-values suggest that continuing would result in negative expected reward, just as humans throw away hopeless hands in real life poker. Similarly, humans raise only when they are confident that they hold either a really strong hand or they are confident that the opponent is bluffing or they know that by raising they could make the opponent fold, and DQN behaves in a similar way where its more risk-averse and avoids over-raising. DQNs call-heavy response pattern shows that it has learned a conservative, information-gathering strategy rather than an exploitative one. When we split the responses per game phase, we can see, from Figure \ref{fig: DQN Reactions PreFlop Threshold-based}, that in the pre-flop stage, DQN still behaves in a information-gathering strategy because the public card has not been revealed and uncertainty is high. Calling is a natural middle-ground because it keeps options open without risking too much. Folding would waste potential equity, as even weak hands can sometimes improve once the public card is revealed due to forming a pair for example. However, once the public card is revealed there is not a lot of additional information to be gained, and DQNs primary response becomes folding. This is similar to how humans play poker in that they fold more in post flop because bluffs are harder to spot with community cards being revealed. DQN shifts to a much more conservative strategy post-flop, preferring to cut losses (fold) over information-gathering (call).

\section{Conclusions and Further Research}\label{Conclusions}

We explore bluffing behavior in Leduc Hold'em by training and evaluating two fundamentally different agents, namely, reaction based Deep Q-Networks (DQN) versus  forward-looking game-theoretic Counterfactual Regret Minimization (CFR). We find that both agents are capable of bluffing.

%, but do so in different circumstances, and respond differently to being bluffed. 

We have trained and evaluated DQN and CFR.
%, agents belonging to different paradigms, against each other which required some modification to the CFR algorithm. 
CFR slightly dominated DQN in win rate which is to be expected since CFR computes the Nash-theoretic equilibrium, which is guaranteed to be unexploitable, on average. 
%rmation settings, and considering that the assumptions of the training procedure of DQN were violated it still held up against CFR and did not get over-dominated. 
%
Interestingly, %What is particularly interesting is that 
both agents exhibited bluffing behavior even though they are based on different principles. Neither agent had been taught deception and bluffing explicitly but bluffing emerged because of the training paradigms, rules of the game and because of each other. CFR attempted  more bluffs than DQN but both ended up having roughly the same bluffing success rate.
Looking at the bluffing styles of both agents, CFR preferred to use mid-strength hands to bluff, while DQN used more low-strength hands.

For the reactions of both agents to each others bluffing attempts,  both followed the same patterns of  being exploratory and risk-minimizing. In  the pre-flop phase they remained exploratory while in the post-flop phase they became more conservative once most of the possible information about that round was gathered. 
Both agents  developed a  similar bluffing and response pattern, %even though they were not explicitly taught about bluffing nor deception (nor how to respond). 
which is  surprising since they are  based on different paradigms. Despite this difference,  they developed similar behavior that is  closely linked to how humans play poker.
%in the real world, specifically, bluffing and responses to it. 

\subsection{Limitations and Further Research}\label{Conclusion Limitations and Further Research}

%While this paper provides meaningful insights into the bluffing behavior of two algorithmic agents, several limitations should be acknowledged and potential future directions will be presented.

This work has the following limitations.
First, all experiments were conducted in the 52-card version of  Leduc Hold'em,  a simplified version of real Texas Hold'em regarding the number of players, number of public cards, number of private cards, and betting amounts. It would be interesting to see when more than two players are present, if the agents target specifically some agents with the bluffs, perhaps the weakest links, or do they attempt to bluff everyone equally. 
Additionally, in No Limit poker it would  be interesting to see how the agents decide the betting amount when bluffing.
%, that is, do they go all in, do they always bet a certain amount when attempting to bluff, etc., and are the opponents able to recognize these patterns if they exist.  

Second, due to practical resource limitations, especially in regard to training time and computational power, only a finite number of training episodes and evaluation games could be executed. It is possible that more training episodes or alternative configurations of the hyperparameters could yield stronger or more refined agents and behaviors. 
%It could also be possible that both agents would converge given more training. 
Potential future work could try to run a hyperparameter sweep \cite{wang2019hyper} to make
%to look into whether modifying certain hyperparameters would make 
the agents more stable or perhaps make the agents bluff more/less. 

Third, the DQN implementation used a standard fully connected feedforward neural network. More advanced architectures could have allowed better information tracking and sequential reasoning. Similarly, the CFR agent was based on tabular updates rather than function approximation which limited its ability to generalize across similar states. Future work could look into using different versions of these algorithms and seeing how they perform. %Additionally, it would also be interesting to use completely different algorithms such as A2C, NFSP, etc.

%Fourth, although the two bluff detection methods that were used agreed on the general patterns, the raw numbers could be slightly off due to the possibility of noise being captured as bluffs or certain patterns not being recognized as bluffs. A potential improvement would be to train a bluffing classifier and see how it would classify bluffs. 

Last, the paper did not compare the agents behaviors to those of human players. While this was beyond the scope of this paper, such a comparison could provide a valuable frame of reference for evaluating the realism and interpretability of bluffing strategies learned by the agents.

%Acknowledging these limitations is important in contextualizing the scope and generalization of the findings. Nonetheless, the controlled setup allowed for a focused and interpretable investigation into the emergence and analysis of bluffing in multi-agent settings.

\bibliographystyle{splncs04}
\bibliography{thesis}

%\appendix
%appendices here --- if any

\end{document}